# Extracting Noun Phrases from Large-Scale Texts: A Hybrid Approach and Its Automatic Evaluation

**Kuang-hua Chen and Hsin-Hsi Chen**

Department of Computer Science and Information Engineering
National Taiwan University
Taipei, Taiwan, R.O.C.
Internet: hh_chen@csie.ntu.edu.tw

**Abstract**

To acquire noun phrases from running texts is useful for many applications, such as word grouping, terminology indexing, *etc*. The reported literatures adopt pure probabilistic approach, or pure rule-based noun phrases grammar to tackle this problem. In this paper, we apply a probabilistic chunker to deciding the implicit boundaries of constituents and utilize the linguistic knowledge to extract the noun phrases by a finite state mechanism. The test texts are SUSANNE Corpus and the results are evaluated by comparing the parse field of SUSANNE Corpus automatically. The results of this preliminary experiment are encouraging.

## 1. Introduction

From the cognitive point of view, human being must recognize, learn and understand the entities or concepts (concrete or abstract) in the texts for natural language comprehension. These entities or concepts are usually described by noun phrases. The evidences from the language learning of children also show the belief (Snow and Ferguson, 1977). Therefore, if we can grasp the noun phases of the texts, we will understand the texts to some extent. This consideration is also captured by theories of discourse analysis, such as Discourse Representation Theory (Kamp, 1981).

Traditionally, to make out the noun phrases in a text means to parse the text and to resolve the attachment relations among the constituents. However, parsing the text completely is very difficult, since various ambiguities cannot be resolved solely by syntactic or semantic information. Do we really need to fully parse the texts in every application? Some researchers apply shallow or partial parsers (Smadja, 1991; Hindle, 1990) to acquiring specific patterns from texts. These tell us that it is not necessary to completely parse the texts for some applications.

This paper will propose a probabilistic partial parser and incorporate linguistic knowledge to extract noun phrases. The partial parser is motivated by an intuition (Abney, 1991):

(1) When we read a sentence, we read it chunk by chunk.

Abney uses two level grammar rules to implement the parser through pure LR parsing technique. The first level grammar rule takes care of the chunking process. The second level grammar rule tackles the attachment problems among chunks. Historically, our statistics-based partial parser is called *chunker*. The chunker receives tagged texts and outputs a linear chunk sequences. We assign a syntactic head and a semantic head to each chunk. Then, we extract the plausible maximal noun phrases according to the information of syntactic head and semantic head, and a finite state mechanism with only 8 states.

Section 2 will give a brief review of the works for the acquisition of noun phrases. Section 3 will describe the language model for chunker. Section 4 will specify how to apply linguistic knowledge to assigning heads to each chunk. Section 5 will list the experimental results of chunker. Following Section 5, Section 6 will give the performance of our work on the retrieval of noun phrases. The possible extensions of the proposed work will be discussed in Section 7. Section 8 will conclude the remarks.

## 2. Previous Works

Church (1988) proposes a part of speech tagger and a simple noun phrase extractor. His noun phrase extractor brackets the noun phrases of input tagged texts according to two probability matrices: one is starting noun phrase matrix; the other is ending noun phrase matrix. The methodology is a simple version of Garside and Leech's probabilistic parser (1985). Church lists a sample text in the Appendix of his paper to show the performance of his work. It demonstrates only 5 out of 248 noun phrases are omitted. Because the tested text is too small to assess the results, the experiment for large volume of texts is needed.

Bourigault (1992) reports a tool, *LEXTER*, for extracting terminologies from texts. *LEXTER* triggers two-stage processing: 1) *analysis* (by identification of frontiers), which extracts the maximal-length noun phrase; 2) *parsing* (the maximal-length noun phrases), which, furthermore, acquires the terminology embedded in the noun phrases. Bourigault declares the *LEXTER* extracts 95% maximal-length noun phrases, that is, 43500 out of 46000 from test corpus. The result is validated by an expert. However, the precision is not reported in the Boruigault's paper.

Voutilainen (1993) announces *NPtool* for acquisition of maximal-length noun phrases. NPtool applies two finite state mechanisms (one is NP-hostile; the other is NP-friendly) to the task. The two mechanisms produce two NP sets and any NP candidate with at least one occurrence in both sets will be labeled as the "ok" NP. The reported recall is 98.5-100% and the precision is 95-98% validated manually by some 20000 words. But from the sample text listed in Appendix of his paper, the recall is about 85% and we can find some inconsistencies among these extracted noun phrases.

## 3. Language Model

Parsing can be viewed as optimizing. Suppose an $n$-word sentence, $w_1, w_2, ..., w_n$ (including punctuation marks), the parsing task is to find a parsing tree $T$, such that $P(T|w_1, w_2, ..., w_n)$ has the maximal probability. We define $T$ here to be a sequence of chunks, $c_1, c_2, ..., c_m$, and each $c_i$ $(0 < i \leq m)$ contains one or more words $w_j$ $(0 < j \leq n)$. For example, the sentence "parsing can be viewed as optimization." consists of 7 words. Its one possible parsing result under our demand is:

(2)  [parsing]  [can be viewed]  [as optimization] [.]
       $c_1$         $c_2$              $c_3$         $c_4$

Now, the parsing task is to find the best chunk sequence, $C^*$, such that

(3)  $C^* = \underset{C_i}{\operatorname{argmax}} P(C_i|w_1^n)$

The $C_i$ is one possible chunk sequence, $c_1, c_2, ..., c_{m_i}$, where $m_i$ is the number of chunks of the possible chunk sequence. To chunk raw text without other information is very difficult, since the word patterns are many millions. Therefore, we apply a tagger to preprocessing the raw texts and give each word a unique part of speech. That is, for an $n$-word sentence, $w_1, w_2, ..., w_n$ (including punctuation marks), we assign part of speeches $t_1, t_2, ..., t_n$ to the respective words. Now the real working model is:

(4)  $C^* = \underset{C_i}{\operatorname{argmax}} P(C_i|t_1^n)$

Using bi-gram language model, we then reduce $P(C_i|t_1, t_2, ..., t_n)$ as (5),

(5)  $P(C_i|t_1^n) = P_i(c_1^n|t_1^n)$
$\cong \prod_{k=1}^{m_i} P_i(c_k|c_{k-1},t_1^n) \times P_i(c_k|t_1^n)$
$\cong \prod_{k=1}^{m_i} P_i(c_k|c_{k-1}) \times P_i(c_k)$

where $P_i(\cdot)$ denotes the probability for the $i$'th chunk sequence and $c_0$ denotes the beginning mark of a sentence. Following (5), formula (4) becomes

(6)  $\underset{C_i}{\operatorname{argmax}} P(C_i|t_1^n)$
$\cong \underset{C_i}{\operatorname{argmax}} \prod_{k=1}^{m_i} P_i(c_k|c_{k-1}) \times P_i(c_k)$
$= \underset{C_i}{\operatorname{argmax}} \sum_{k=1}^{m_i} [\log(P_i(c_k|c_{k-1})) + \log(P_i(c_k))]$

In order to make the expression (6) match the intuition of human being, namely, 1) the scoring metrics are all positive, 2) large value means high score, and 3) the scores are between 0 and 1, we define a score function $S(P(\cdot))$ shown as (7).

(7)  $S(P(\cdot)) = 0$   when $P(\cdot) = 0$;
     $S(P(\cdot)) = 1.0/(1.0+\text{ABS}(\log(P(\cdot))))$   o/w.

We then rewrite (6) as (8).

(8)  $C^* = \underset{C_i}{\operatorname{argmax}} P(C_i|t_1^n)$
$\cong \underset{C_i}{\operatorname{argmax}} \prod_{k=1}^{m_i} P_i(c_k|c_{k-1}) \times P_i(c_k)$
$= \underset{C_i}{\operatorname{argmax}} \sum_{k=1}^{m_i} [\log(P_i(c_k|c_{k-1})) + \log(P_i(c_k))]$
$= \underset{C_i}{\operatorname{argmax}} \sum_{k=1}^{m_i} [S(P_i(c_k|c_{k-1})) + S(P_i(c_k))]$

The final language model is to find a chunk sequence $C^*$, which satisfies the expression (8).

Dynamic programming shown in (9) is used to find the best chunk sequence. The *score[i]* denotes the score of position $i$. The words between position *pre[i]* and position $i$ form the best chunk from the viewpoint of position $i$. The *dscore(ci)* is the score of the probability

$P(c_i)$ and the *cscore($c_i$|$c_{i-1}$)* is the score of the probability $P(c_i|c_{i-1})$. These scores are collected from the training corpus, SUSANNE corpus (Sampson, 1993; Sampson, 1994). The details will be touched on in Section 5.

---

(9) Algorithm
  input : word sequence $w_1, w_2, ..., w_n$, and
      the corresponding POS sequence $t_1, t_2, ..., t_n$
  output : a sequence of chunks $c_1, c_2, ..., c_m$
  1. score[0] = 0;
     pre[0] = 0;
  2. for (i = 1; i<n+1; i++) do 3 and 4;
  3. j*= $\maxarg_{0 \leq j < i}$(score[pre[j]]+dscore($c_j$)+cscore($c_j$|$c_{j-1}$));
       where  $c_j = t_{j+1}, ..., t_i$;
            $c_{j-1} = t_{pre[j]+1}, ..., t_j$;
  4. score[i]=score[pre[j*]]+dscore($c_{j*}$)+cscore($c_{j*}$|$c_{j*-1}$);
     pre[i] = j*;
  5. for (i=n; i>0; i=pre[i]) do
     output the word $w_{pre[i]+1}, ..., w_i$ to form a chunk;

---

## 4. Linguistic Knowledge

In order to assign a head to each chunk, we first define priorities of POSes. X'-theory (Sells, 1985) has defined the X'-equivalences shown as Table 1.

**Table 1. X'-Equivalences**

| X | X' | X" |
|---|---|---|
| N | N' | NP |
| V | V' | VP |
| A | A' | AP |
| P | P' | PP |
| INFL | S (I') | S' (IP) |

Table 1 defines five different phrasal structures and the hierarchical structures. The heads of these phrasal structures are the first level of X'-Equivalences, that is, X. The other grammatical constituents function as the specifiers or modifiers, that is, they are accompanying words not core words. Following this line, we define the primary priority of POS listed in Table 1.

(10) Primary POS priority[1] : V > N > A > P

In order to extract the exact head, we further define Secondary POS priority among the 134 POSes defined in LOB corpus (Johansson, 1986).

(11) Secondary POS priority is a linear precedence relationship within the primary priorities for coarse POSes

---

[1] We do not consider the INFL, since our model will not touch on this structure.

For example, LOB corpus defines four kinds of verbial words under the coarse POS V: VB∗, DO∗, BE∗ and HV∗[2]. The secondary priority within the coarse POS V is:

(12)  VB∗ > HV∗ > DO∗ > BE∗

Furthermore, we define the semantic head and the syntactic head (Abney, 1991).

(13) Semantic head is the head of a phrase according to the semantic usage; but syntactic head is the head based on the grammatical relations.

Both the syntactic head and the semantic head are useful in extracting noun phrases. For example, if the semantic head of a chunk is the noun and the syntactic one is the preposition, it would be a prepositional phrase. Therefore, it can be connected to the previous noun chunk to form a new noun phrase. In some case, we will find some chunks contain only one word, called *one-word* chunks. They maybe contain a conjunction, e.g., that. Therefore, the syntactic head and the semantic head of *one-word* chunks are the word itself.

Following these definitions, we extract the noun phrases by procedure (14):

(14) (a) Tag the input sentences.
    (b) Partition the tagged sentences into chunks by using a probabilistic partial parser.
    (c) Decide the syntactic head and the semantic head of each chunk.
    (d) According to the syntactic and the semantic heads, extract noun phrase from these chunks and connect as many noun phrases as possible by a finite state mechanism.

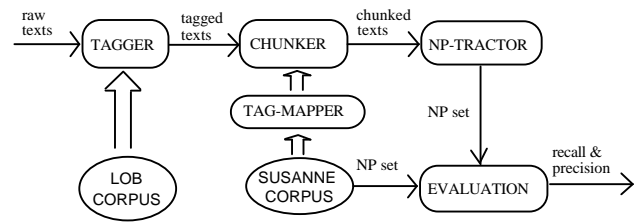

Figure 1. The Noun Phrases Extraction Procedure

Figure 1 shows the procedure. The input raw texts will be assigned POSes to each word and then pipelined into a chunker. The tag sets of LOB and SUSANNE are different. Since the tag set of SUSANNE corpus is subsumed by the tag set of LOB corpus, a TAG-MAPPER is used to map tags of SUSANNE corpus to those of LOB corpus. The chunker will output a sequence of chunks. Finally, a finite state NP-TRACTOR will extract NPs. Figure 2 shows the finite state mechanism used in our work.

---

[2] Asterisk ∗ denotes wildcard. Therefore, VB∗ represents VB (verb, base form), VBD (verb, preterite), VBG (present participle), VBN (past participle) and VBZ (3rd singular form of verb).

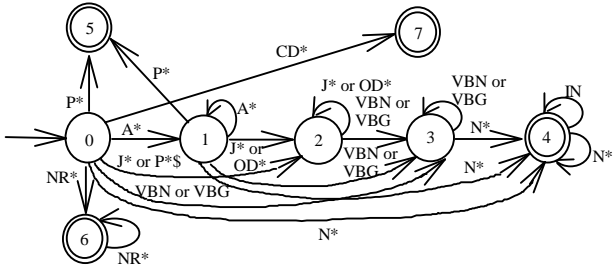

Figure 2. The Finite State Machine for Noun Phrases

The symbols in Figure 2 are tags of LOB corpus. N* denotes nous; P* denotes pronouns; J* denotes adjectives; A* denotes quantifiers, qualifiers and determiners; IN denotes prepositions; CD* denotes cardinals; OD* denotes ordinals, and NR* denotes adverbial nouns. Asterisk * denotes a wildcard. For convenience, some constraints, such as syntactic and semantic head checking, are not shown in Figure 2.

## 5. First Stage of Experiments

Following the procedures depicted in Figure 1, we should train a chunker firstly. This is done by using the SUSANNE Corpus (Sampson, 1993; Sampson, 1994) as the training texts. The SUSANNE Corpus is a modified and condensed version of Brown Corpus (Francis and Kucera, 1979). It only contains the 1/10 of Brown Corpus, but involves more information than Brown Corpus. The Corpus consists of four kinds of texts: 1) A: press reportage; 2) G: belles letters, biography, memoirs; 3) J: learned writing; and 4) N: adventure and Western fiction. The Categories of A, G, J and N are named from respective categories of the Brown Corpus. Each Category consists of 16 files and each file contains about 2000 words.

The following shows a snapshot of SUSANNE Corpus.

```
G01:0010a - YB    <minbrk>      -            [Oh.Oh]
G01:0010b - JJ    NORTHERN      northern     [O[S[Np:s.
G01:0010c - NN2   liberals      liberal      .Np:s]
G01:0010d - VBR   are           be           [Vab.Vab]
G01:0010e - AT    the           the          [Np:e.
G01:0010f - JB    chief         chief        .
G01:0010g - NN2   supporters    supporter    .
G01:0010h - IO    of            of           [Po.
G01:0010i - JJ    civil         civil        [Np.
G01:0010j - NN2   rights        right        .Np]
G01:0020a - CC    and           and          [Po+.
G01:0020b - IO    of            of           .
G01:0020c - NN1u  integration   integration  .Po+]Po]Np:e]S]
G01:0020d - YF    +.            -            .
```

Table 2 lists basic statistics of SUSANNE Corpus.

**Table 2. The Overview of SUSANNE Corpus**

| Categories | Files | Paragraphs | Sentences | Words |
|---|---|---|---|---|
| A | 16 | 767 | 1445 | 37180 |
| G | 16 | 280 | 1554 | 37583 |
| J | 16 | 197 | 1353 | 36554 |
| N | 16 | 723 | 2568 | 38736 |
| Total | 64 | 1967 | 6920 | 150053 |

In order to avoid the errors introduced by tagger, the SUSANNE corpus is used as the training and testing texts. Note the tags of SUSANNE corpus are mapped to LOB corpus. The 3/4 of texts of each categories of SUSANNE Corpus are both for training the chunker and testing the chunker (inside test). The rest texts are only for testing (outside test). Every tree structure contained in the parse field is extracted to form a potential chunk grammar and the adjacent tree structures are also extracted to form a potential context chunk grammar. After the training process, total 10937 chunk grammar rules associated with different scores and 37198 context chunk grammar rules are extracted. These chunk grammar rules are used in the chunking process.

Table 3 lists the time taken for processing SUSANNE corpus. This experiment is executed on the Sun Sparc 10, model 30 workstation. T denotes time, W word, C chunk, and S sentence. Therefore, T/W means the time taken to process a word on average.

**Table 3. The Processing Time**

|  | T/W | T/C | T/S |
|---|---|---|---|
| A | 0.00295 | 0.0071 | 0.0758 |
| G | 0.00283 | 0.0069 | 0.0685 |
| J | 0.00275 | 0.0073 | 0.0743 |
| N | 0.00309 | 0.0066 | 0.0467 |
| Av. | 0.00291 | 0.0070 | 0.0663 |

According to Table 3, to process a word needs 0.00291 seconds on average. To process all SUSANNE corpus needs about 436 seconds, or 7.27 minutes.

In order to evaluate the performance of our chunker, we compare the results of our chunker with the denotation made by the SUSANNE Corpus. This comparison is based on the following criterion:

(15) The content of each chunk should be dominated by one non-terminal node in SUSANNE parse field.

This criterion is based on an observation that each non-terminal node has a chance to dominate a chunk.

Table 4 is the experimental results of testing the SUSANNE Corpus according to the specified criterion. As usual, the symbol C denotes chunk and S denotes sentence.

**Table 4. Experimental Results**

| Cat. | TEST | OUTSIDE TEST | | INSIDE TEST | |
|---|---|---|---|---|---|
| | | C | S | C | S |
| A | # of correct | 4866 | 380 | 10480 | 1022 |
| | # of incorrect | 40 | 14 | 84 | 29 |
| | total # | 4906 | 394 | 10564 | 1051 |
| | correct rate | 0.99 | 0.96 | 0.99 | 0.97 |
| G | # of correct | 4748 | 355 | 10293 | 1130 |
| | # of incorrect | 153 | 32 | 133 | 37 |
| | total # | 4901 | 387 | 10426 | 1167 |
| | correct rate | 0.97 | 0.92 | 0.99 | 0.97 |
| J | # of correct | 4335 | 283 | 9193 | 1032 |
| | # of incorrect | 170 | 15 | 88 | 23 |
| | total # | 4505 | 298 | 9281 | 1055 |
| | correct rate | 0.96 | 0.95 | 0.99 | 0.98 |
| N | # of correct | 5163 | 536 | 12717 | 1906 |
| | # of incorrect | 79 | 42 | 172 | 84 |
| | total # | 5242 | 578 | 12889 | 1990 |
| | correct rate | 0.98 | 0.93 | 0.99 | 0.96 |
| Av. | # of correct | 19112 | 1554 | 42683 | 5090 |
| | # of incorrect | 442 | 103 | 477 | 173 |
| | total # | 19554 | 1657 | 43160 | 5263 |
| | correct rate | 0.98 | 0.94 | 0.99 | 0.97 |

Table 4 shows the chunker has more than 98% chunk correct rate and 94% sentence correct rate in outside test, and 99% chunk correct rate and 97% sentence correct rate in inside test. Note that once a chunk is mischopped, the sentence is also mischopped. Therefore, sentence correct rate is always less than chunk correct rate. Figure 3 gives a direct view of the correct rate of this chunker.

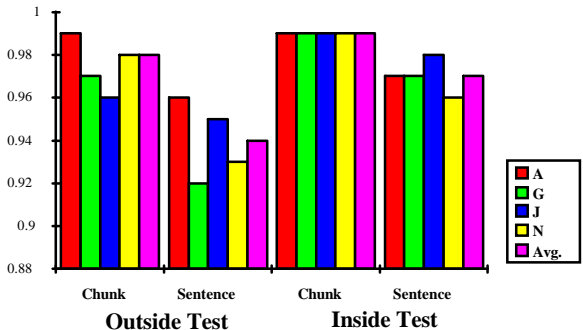

Figure 3. The Correct Rate of Experiments

## 6. Acquisition of Noun Phrases

We employ the SUSANNE Corpus as test corpus. Since the SUSANNE Corpus is a parsed corpus, we may use it as criteria for evaluation. The volume of test texts is around 150,000 words including punctuation marks. The time needed from inputting texts of SUSANNE Corpus to outputting the extracted noun phrases is listed in Table 5. Comparing with Table 3, the time of combining chunks to form the candidate noun phrases is not significant.

**Table 5. Time for Acquisition of Noun Phrases**

| | Words | Time (sec.) | Time/Word |
|---|---|---|---|
| A | 37180 | 112.32 | 0.00302 |
| G | 37583 | 108.80 | 0.00289 |
| J | 36554 | 103.04 | 0.00282 |
| N | 38736 | 122.72 | 0.00317 |
| Total | 150053 | 446.88 | 0.00298 |

The evaluation is based on two metrics: precision and recall. Precision means the correct rate of what the system gets. Recall indicates the extent to which the real noun phrases retrieved from texts against the real noun phrases contained in the texts. Table 6 describes how to calculate these metrics.

**Table 6. Contingency Table for Evaluation**

| | | SUSANNE | |
|---|---|---|---|
| | | NP | non-NP |
| System | NP | a | b |
| | non-NP | c | - |

The rows of "System" indicate our NP-TRACTOR thinks the candidate as an NP or not an NP; the columns of "SUSANNE" indicate SUSANNE Corpus takes the candidate as an NP or not an NP. Following Table 6, we will calculate precision and recall shown as (16).

(16) Precision = a/(a+b) * 100%
     Recall = a/(a+c) * 100%

To calculate the precision and the recall based on the parse field of SUSANNE Corpus is not so straightforward at the first glance. For example, (17)[3] itself is a noun phrse but it contains four noun phrases. A tool for extracting noun phrases should output what kind of and how many noun phrases, when it processes the texts like (17). Three kinds of noun phrases (maximal noun phrases, minimal noun phrases and ordinary noun phrases) are defined first. Maximal noun phrases are those noun phrases which are not contained in other noun phrases. In contrast, minimal noun phrases do not contain any other noun phrases.

---

[3] This example is taken from N06:0280d-N06:0290d, Susanne Corpus (N06 means file N06, 0280 and 0290 are the original line numbers in Brown Corpus. Recall that the Susanne Corpus is a modified and reduced version of Brown Corpus).

Apparently, a noun phrase may be both a maximal noun phrase and a minimal noun phrase. Ordinary noun phrases are noun phrases with no restrictions. Take (17) as an example. It has three minimal noun phrases, one maximal noun phrases and five ordinary noun phrases. In general, a noun-phrase extractor forms the front end of other applications, e.g., acquisition of verb subcategorization frames. Under this consideration, it is not appropriate to taking (17) as a whole to form a noun phrase. Our system will extract two noun phrases from (17), "a black badge of frayed respectability" and "his neck".

(17) [[[a black badge] of [frayed respectability]] that ought never to have left [his neck]]

We calculate the numbers of maximal noun phrases, minimal noun phrases and ordinary noun phrases denoted in SUSANNE Corpus, respectively and compare these numbers with the number of noun phrases extracted by our system.

Table 7 lists the number of ordinary noun phrases (NP), maximal noun phrases (MNP), minimal noun phrases (mNP) in SUSANNE Corpus. MmNP denotes the maximal noun phrases which are also the minimal noun phrases. On average, a maximal noun phrase subsumes 1.61 ordinary noun phrases and 1.09 minimal noun phrases.

**Table 7. The Number of Noun Phrases in Corpus**

|  | NP | MNP | mNP | MmNP | $\frac{NP}{MNP}$ | $\frac{mNP}{MNP}$ |
|---|---|---|---|---|---|---|
| A | 10063 | 5614 | 6503 | 3207 | 1.79 | 1.16 |
| G | 9221 | 5451 | 6143 | 3226 | 1.69 | 1.13 |
| J | 8696 | 4568 | 5200 | 2241 | 1.90 | 1.14 |
| N | 9851 | 7895 | 7908 | 5993 | 1.25 | 1.00 |
| Total | 37831 | 23528 | 25754 | 14667 | 1.61 | 1.09 |

To calculate the precision, we examine the extracted noun phrases (ENP) and judge the correctness by the SUSANNE Corpus. The CNP denotes the correct ordinary noun phrases, CMNP the correct maximal noun phrases, CmNP correct minimal noun phrases and CMmNP the correct maximal noun phrases which are also the minimal noun phrases. The results are itemized in Table 8. The average precision is 95%.

**Table 8. Precision of Our System**

|  | ENP | CNP | CMNP | CmNP | CMmNP | Precision |
|---|---|---|---|---|---|---|
| A | 8011 | 7660 | 3709 | 4348 | 3047 | 0.96 |
| G | 7431 | 6943 | 3626 | 4366 | 3028 | 0.93 |
| J | 6457 | 5958 | 2701 | 3134 | 2005 | 0.92 |
| N | 8861 | 8559 | 6319 | 6637 | 5808 | 0.97 |
| Total | 30760 | 29120 | 16355 | 18485 | 13888 | 0.95 |

Here, the computation of recall is ambiguous to some extent. Comparing columns CMNP and CmNP in Table 8 with columns MNP and mNP in Table 7, 70% of MNP and 72% of mNP in SUSANNE Corpus are extracted. In addition, 95% of MmNP is extracted by our system. It means the recall for extracting noun phrases that exist independently in SUSANNE Corpus is 95%. What types of noun phrases are extracted are heavily dependent on what applications we will follow. We will discuss this point in Section 7. Therefore, the real number of the applicable noun phrases in the Corpus is not known. The number should be between the number of NPs and that of MNPs. In the original design for NP-TRACTOR, a maximal noun phrase which contains clauses or prepositional phrases with prepositions other than "of" is not considered as an extracted unit. As the result, the number of such kinds of applicable noun phrases (ANPs) form the basis to calculate recall. These numbers are listed in Table 9 and the corresponding recalls are also shown.

**Table 9. The limitation of Values for Recall**

|  | ANP | CNP | Recall |
|---|---|---|---|
| A | 7873 | 7660 | 0.97 |
| G | 7199 | 6943 | 0.96 |
| J | 6278 | 5958 | 0.95 |
| N | 8793 | 8559 | 0.97 |
| Av. | 30143 | 29120 | 0.96 |

The automatic validation of the experimental results gives us an estimated recall. Appendix provides a sample text and the extracted noun phrases. Interested readers could examine the sample text and calculate recall and precision for a comparison.

## 7. Applications

Identification of noun phrases in texts is useful for many applications. Anaphora resolution (Hirst, 1981) is to resolve the relationship of the noun phrases, namely, what the antecedent of a noun phrase is. The extracted noun phrases can form the set of possible candidates (or universal in the terminology of discourse representation theory). For acquisition of verb subcategorization frames, to bracket the noun phrases in the texts is indispensable. It can help us to find the boundary of the subject, the object and the prepositional phrase. We would use the acquired noun phrases for an application of adjective grouping. The extracted noun phrases may contain adjectives which pre-modify the head noun. We then utilize the similarity of head nouns to group the adjectives. In addition, we may give the head noun a semantic tag, such as Roget's Thesaurus provides, and then analyze the adjectives. To automatically produce the index of a book, we would extract the noun phrases contained in the book, calculate the inverse document frequency (IDF) and their term frequency (TF) (Salton, 1991), and screen out the implausible terms.

These applications also have impacts on identifying noun phrases. For applications like anaphora resolution and acquisition of verb subcategorization frames, the maximal noun phrases are not suitable. For applications like grouping adjectives and automatic book indexing, some kinds of maximal noun phrases, such as noun phrases postmodified by "of" prepositional phrases, are suitable; but some are not, e.g., noun phrases modified by relative clauses.

## 8. Concluding Remarks

The difficulty of this work is how to extract the real maximal noun phrases. If we cannot decide the prepositional phrase "over a husband eyes" is licensed by the verb "pull", we will not know "the wool" and "a husband eyes" are two noun phrases or form a noun pharse combined by the preposition "over".

(18) to pull the wool over a husband eyes
     to sell the books of my uncle

In contrast, the noun phrase "the books of my uncle" is so called maximal noun phrase in current context. As the result, we conclude that if we do not resolve PP-attachment problem (Hindle and Rooth, 1993), to the expected extent, we will not extract the maximal noun phrases. In our work, the probabilistic chunker decides the implicit boundaries between words and the NP-TRACTOR connects the adjacent noun chunks. When a noun chunk is followed by a preposition chunk, we do not connect the two chunks except the preposition chunk is led by "of" preposition.

Comparing with other works, our results are evaluated by a parsed corpus automatically and show the high precision. Although we do not point out the exact recall, we provide estimated values. The testing scale is large enough (about 150,000 words). In contrast, Church (1988) tests a text and extracts the simple noun phrases only. Bourigault's work (1992) is evaluated manually, and dose not report the precision. Hence, the real performance is not known. The work executed by Voutilainen (1993) is more complex than our work. The input text first is morphologizied, then parsed by constraint grammar, analyzed by two different noun phrases grammar and finally extracted by the occurrences. Like other works, Voutilainen's work is also evaluated manually.

In this paper, we propose a language model to chunk texts. The simple but effective chunker could be seen as a linear structure parser, and could be applied to many applications. A method is presented to extract the noun phrases. Most importantly, the relations of maximal noun phrases, minimal noun phrases, ordinary noun phrases and applicable noun phrases are distinguished in this work. Their impacts on the subsequent applications are also addressed. In addition, automatic evaluation provides a fair basis and does not involve human costs. The experimental results show that this parser is a useful tool for further research on large volume of real texts.

## Acknowledgements
We are grateful to Dr. Geoffrey Sampson for his kindly providing SUSANNE Corpus and the details of tag set to us.

## Appendix

For demonstration, we list a sample text quoted from N18:0010a-N18:0250e, SUSANNE Corpus. The extracted noun phrases are bracketed. We could compute the precision and the recall from the text as a reference and compare the gap with the experimental results itemized in Section 6. In actual, the result shows that the system has high precision and recall for the text.

[ Too_QL many_AP people_NNS ] think_VB that_CS [ the_ATI primary_JJ purpose_NN of_IN a_AT higher_JJR education_NN ] is_BEZ to_TO help_VB [ you_PP2 ] make_VB [ a_AT living_NN ] +;_; this_DT is_BEZ not_XNOT so_RB +,_, for_CS [ education_NN ] offers_VBZ [ all_ABN kinds_NNS of_IN dividends_NNS ] +,_, including_IN how_WRB to_TO pull_VB [ the_ATI wool_NN ] over_IN [ a_AT husband_NN eyes_NNS ] while_CS [ you_PP2 ] are_BER having_HVG [ an_AT affair_NN ] with_IN [ his_PP$ wife_NN ] ._. If_CS [ it_PP3 ] were_BED not_XNOT for_IN [ an_AT old_JJ professor_NPT ] who_WPR made_VBD [ me_PP1O ] read_VB [ the_ATI classics_NN ] [ I_PP1A ] would_MD have_HV been_BEN stymied_VBN on_IN what_WDT to_TO do_DO +,_, and_CC now_RN [ I_PP1A ] understand_VB why_WRB [ they_PP3AS ] are_BER [ classics_NN ] +;_; those_DTS who_WPR wrote_VBD [ them_PP3OS ] knew_VBD [ people_NNS ] and_CC what_WDT made_VBD [ people_NNS ] tick_VB ._. [ I_PP1A ] worked_VBD for_IN [ my_PP$ Uncle_NPT ] (_( [ +an_AT Uncle_NPT by_IN marriage_NN ] so_RB [ you_PP2 ] will_MD not_XNOT think_VB this_DT has_HVZ [ a_AT mild_JJ undercurrent_NN of_IN incest_NN ] +)_) who_WPR ran_VBD [ one_CD1 of_IN those_DTS antique_JJ shops_NNS ] in_IN [ New_JJ Orleans_NP ] Vieux_&FW Carre_&FW +,_, [ the_ATI old_JJ French_JJ Quarter_NPL ] ._. [ The_ATI arrangement_NN ] [ I_PP1A ] had_HVD with_IN [ him_PP3O ] was_BEDZ to_TO work_VB [ four_CD hours_NRS ] [ a_AT day_NR ] ._. [ The_ATI rest_NN of_IN the_ATI time_NR ] [ I_PP1A ] devoted_VBD to_IN painting_VBG or_CC to_IN those_DTS [ other_JJB activities_NNS ] [ a_AT young_JJ and_CC healthy_JJ man_NN ] just_RB out_IN of_IN [ college_NN ] finds_VBZ interesting_JJ ._. [ I_PP1A ] had_HVD [ a_AT one-room_JJ studio_NN ] which_WDTR overlooked_VBD [ an_AT ancient_JJ courtyard_NN ] filled_VBN with_IN [ flowers_NNS and_CC plants_NNS ] +,_, blooming_VBG everlastingly_RB in_IN [ the_ATI southern_JJ sun_NN ] ._. [ I_PP1A ] had_HVD come_VBN to_IN [ New_JJ Orleans_NP ] [ two_CD years_NRS ] earlier_RBR after_IN [ graduating_VBG college_NN ] +,_, partly_RB because_CS [ I_PP1A ] loved_VBD [ the_ATI city_NPL ] and_CC partly_RB because_CS there_EX was_BEDZ quite_QL [ a_AT noted_JJ art_NN colony_NN ] there_RN ._. When_CS [ my_PP$ Uncle_NPT ] offered_VBD [ me_PP1O ] [ a_AT part-time_JJ job_NN ] which_WDTR would_MD take_VB [ care_NN ] of_IN [ my_PP$ normal_JJ expenses_NNS ] and_CC give_VB [ me_PP1O ] [ time_NR ] to_TO paint_VB [ I_PP1A ] accepted_VBD ._. [ The_ATI arrangement_NN ] turned_VBD out_RP to_TO be_BE excellent_JJ ._. [ I_PP1A ] loved_VBD [ the_ATI city_NPL ] and_CC [ I_PP1A ] particularly_RB loved_VBD [ the_ATI gaiety_NN and_CC spirit_NN ] of_IN [ Mardi_NR Gras_NR ] ._. [ I_PP1A ] had_HVD seen_VBN [ two_CD of_IN them_PP3OS ] and_CC [ we_PP1AS ] would_MD soon_RB be_BE in_IN another_DT city-wide_JJ +,_, [ joyous_JJ celebration_NN with_IN romance_NN ] in_IN [ the_ATI air_NN ] +;_; and_CC +,_, when_CS [ you_PP2 ] took_VBD [ a_AT walk_NPL ] [ you_PP2 ] never_RB knew_VBD what_WDT [ adventure_NN or_CC pair_NN of_IN sparkling_JJ eyes_NNS ] were_BED waiting_VBG around_IN [ the_ATI next_OD corner_NPL ] ._.